\documentclass[runningheads]{llncs}
\usepackage{graphicx}
\usepackage{amsmath}
\usepackage[titlenumbered,ruled]{algorithm2e}
\usepackage{multirow}
\usepackage{subfig}

\begin{document}
\title{Temporal Continuity Based Unsupervised Learning for Person Re-Identification}
\titlerunning{Temporal Continuity for Person Re-Id}
% If the paper title is too long for the running head, you can set
% an abbreviated paper title here
%
\author{Usman Ali\inst{1} \and
Bayram Bayramli\inst{1} \and
Hongtao Lu\inst{1}}
\authorrunning{U. Ali et al.}
% First names are abbreviated in the running head.
% If there are more than two authors, 'et al.' is used.
%
\institute{Shanghai Jiao Tong University, China \\
\email{\{usmanali,bayram\_bai,htlu\}@sjtu.edu.cn}}
\maketitle              % typeset the header of the contribution
\begin{abstract}
Person re-identification (re-id) aims to match the same person from images taken across multiple cameras. Most existing person re-id methods generally require a large amount of identity labeled data to act as discriminative guideline for representation learning. Difficulty in manually collecting identity labeled data leads to poor adaptability in practical scenarios. To overcome this problem, we propose an unsupervised center-based clustering approach capable of progressively learning and exploiting the underlying re-id discriminative information from temporal continuity within a camera. We call our framework Temporal Continuity based Unsupervised Learning (TCUL). Specifically, TCUL simultaneously does center based clustering of unlabeled (target) dataset and fine-tunes a convolutional neural network (CNN) pre-trained on irrelevant labeled (source) dataset to enhance discriminative capability of the CNN for the target dataset. Furthermore, it exploits temporally continuous nature of images within-camera jointly with spatial similarity of feature maps across-cameras to generate reliable pseudo-labels for training a re-identification model. As the training progresses, number of reliable samples keep on growing adaptively which in turn boosts representation ability of the CNN. Extensive experiments on three large-scale person re-id benchmark datasets are conducted to compare our framework with state-of-the-art techniques, which demonstrate superiority of TCUL over existing methods.

\end{abstract}

\section{Introduction}
Person re-identification (re-id) is an important problem in computer vision that aims to match images of a person captured by different cameras with non-overlapping views \cite{zheng2016person}. It is viewed as a search problem with a goal to retrieve the most relevant images to the top ranks \cite{scalabale_reid}. Many recent works have attracted extensive research efforts to address the re-id problem using deep learning \cite{dml_reid,filter_pair,Hermans2017InDO,Xiao2016LearningDF,Li2018HarmoniousAN,Su2017PoseDrivenDC}. In spite of remarkable advancements achieved by deep learning methods, most existing techniques adopt a supervised learning paradigm to solve the re-id problem \cite{dml_reid,filter_pair}. Hence, these supervised deep models take an assumption of availability of sufficient cross-view identity matching pairs of manually labelled training data for each camera network. This assumption limits the generalizing capability of a re-id model to multiple camera networks due to lack of labelled training samples under a new environment.

As a result, many previous works have addressed this scarcity of labelled training data under a new camera network by focusing on unsupervised or semi-supervised learning \cite{Khan2016UnsupervisedDA,Kodirov2015DictionaryLW,Lisanti2015PersonRB,Deng2018ImageImageDA,Wei2018PersonTG}.  Most of these works typically deal with relatively small datasets; without being able to exploit the potential of deep learning methods that require large-scale datasets for better performance \cite{Khan2016UnsupervisedDA,Kodirov2016PersonRB,Ma2017PersonRB,Ye2017DynamicLG}. More recently, there has been greater emphasis towards clustering and domain adaptation based deep unsupervised methods for person re-id \cite{Fan2018UnsupervisedPR,Peng2016UnsupervisedCT,Li2018AdaptationAR}. However, performance of unsupervised learning approaches is much weaker as compared to supervised models. This is due to nonexistence of cross-view identity labelled matching pairs, which makes it difficult to learn the discriminative representation for recognizing a person under severe changes in visual appearances across different camera networks.

To improve this situation, we propose to use temporal continuity of images within a camera jointly with spatial similarity of feature maps across-cameras to generate reliable pseudo-labels for training a re-id model in a progressive fashion. Temporal information is readily available for images captured by a camera, for example, frame number (Frame ID) of an image in a video sequence is easily obtainable in an unsupervised manner. The idea is that given an image in a camera, other images of the same identity would exist in the temporal vicinity of that image. However, temporal continuity is only effective within one camera; hence, it cannot be utilized to obtain cross-camera matching identity pairs. To this end, we propose a cross-camera center-based pseudo-labeling method to cluster features by their similarity to centers. Centers are initialized by performing a simple clustering task on features extracted from a baseline model pre-trained on irrelevant (source) dataset. An intersection between cross-camera pseudo-labels and within-camera temporal vicinity results in highly reliable pseudo-labels for fine-tuning a re-id model. Features extracted from fine-tuned model are then used to generate a new set of more reliable labels for further fine-tuning in a progressive manner. It is also notable here that different from video-based method \cite{Li2018UnsupervisedPR,Chen2018DeepAL}, our method does not rely on person trajectory obtained by tracking algorithms. Therefore, it is highly applicable to practical scenarios.

In summary, the contributions of this paper are as follows:
\begin{itemize}
\item we propose center-update based pseudo-labeling approach in order to provide cross-view discriminative information for unsupervised person re-id task.
\item we exploit temporal continuity of images within a camera to sample reliable data-points and utilize self-paced progressive learning to train an effective re-id model. To the best of our knowledge, this is the first work to use temporal continuity for person re-id task.
\item extensive experiments on three large-scale datasets indicate the effectiveness of our method with performance on par with state-of-the-art approaches.
\end{itemize}

\section{Related Work}
In this section, we briefly review some recent advances in supervised person re-id and unsupervised person re-id.

\subsection{Person Re-identification}
Most existing works on person re-id are supervised. Classical person re-id models are either based on distance metric learning \cite{Lisanti2014MatchingPA,Hirzer2012RelaxedPL} or rank learning \cite{Paisitkriangkrai2015LearningTR}. Recently, many works have adopted deep learning based approaches to achieve state-of-the-art results in person re-id. We broadly group deep learning models for re-id task into two categories, metric learning and feature representation learning. Metric learning methods learn a similarity metric using image pairs or triplets as inputs \cite{dml_reid,filter_pair}. Li et al. \cite{dml_reid} used a deep siamese network to learn a similarity metric directly from image pixels and Yi et al. \cite{filter_pair} proposed a patch matching based filter pairing network to classify similar and dissimilar images. Cheng et al. \cite{Cheng2016PersonRB} used a part-based approach to first divide features into multiple parts and finally merge all the parts together to calculate triplet loss. On the other hand, feature representation learning methods are more suitable for large-scale datasets which generally use classification model to address re-id task. A pedestrian identity-predicting model is trained to extract features from last layers as discriminative embedding during testing \cite{zheng2016person,Ding2017LetFD,Zheng2017ADL}. Hermans et al. \cite{Hermans2017InDO} employed an efficient variant of triplet loss based on distance among features to learn identity discriminative representation. To address the scarcity of data, some works like \cite{Zheng2017UnlabeledSG,feature_affinity} demonstrated the effectiveness of generative adversarial networks to generate additional samples for re-id training. In this work, we adopt a supervised model trained on source data as our baseline re-id model.

\subsection{Unsupervised Person Re-identification}
Many classical unsupervised learning methods for person re-id employ handcrafted features and give poor performance when compared to supervised methods \cite{Khan2016UnsupervisedDA,Kodirov2015DictionaryLW,Lisanti2015PersonRB,Ye2017DynamicLG}. Recently, we can see some works focusing on unsupervised re-id using deep learning. Li et al. \cite{Li2018UnsupervisedPR} used video-based person trajectory information obtained from tracking algorithms to create pseudo-labels for each camera. Wang et al. \cite{Wang2018TransferableJA} proposed a network which learns to transfer attribute-semantic and identity-discriminative representation across domains via generating pseudo attribute labels. Peng et al. \cite{Peng2016UnsupervisedCT} developed a multi-task dictionary learning framework for cross-dataset representation transfer. A few works also apply generative adversarial networks to adapt representation from one domain to another \cite{Deng2018ImageImageDA,Wei2018PersonTG}. In this work, though focus is not domain adaptation, we use a pre-trained model on labeled source dataset as our baseline model.

To address unsupervised person re-id, generating pseudo-labels for unlabeled dataset is another common practice. Fan et al. proposed a clustering and fine-tuning based method in \cite{Fan2018UnsupervisedPR} known as progressive unsupervised learning (PUL). Although our approach has a similar learning objective as PUL, we distinguish our work in two important aspects. Firstly, in order to generate new pseudo-labels after each iteration, unlike PUL, which uses \textit{k-means}, we propose a feature-based center update for a more stable transition of cluster centers. Secondly, we use temporal continuity to sample reliable data points, in contrast to a distance based selection operator as used by PUL. A distance based selection operator is highly susceptible to faulty assignment, especially in the start of training process. Results in Section \ref{sec:exp} show the superiority of using temporal continuity for sample selection.

\begin{figure*}[t!]
    \centering
    \includegraphics[width=\textwidth]{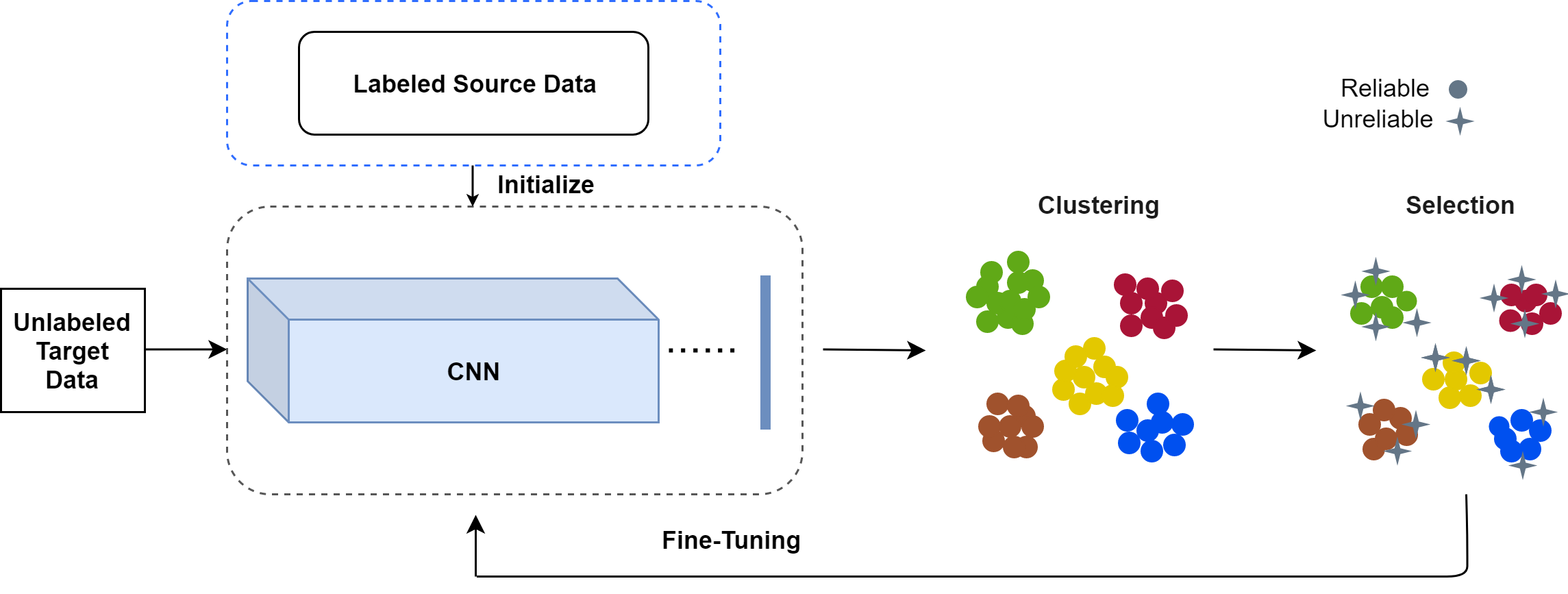}
    \caption{Best viewed in color. Architecture for TCUL. A CNN model initialized on a labeled source dataset is used to extract features from unlabeled target dataset for similarity based clustering. Then reliable samples are selected by exploiting temporal continuity within a camera for fine-tuning the CNN model. Apart from initialization, all other steps are repeated for multiple iterations.}
    \label{fig:main_arch}
\end{figure*}

\section{The Proposed Method}
In this section, we present our proposed method TCUL illustrated in Figure \ref{fig:main_arch}.

\subsection{Parameter Initialization}
Inspired by many recent domain adaptation methods for unsupervised person re-id  \cite{Fan2018UnsupervisedPR,Li2018AdaptationAR,Peng2016UnsupervisedCT}, we initialize our model using an irrelevant labeled dataset (source) which is generated with a completely different camera network and has no identity  overlap with the unlabeled target dataset. Specifically, we train a triplet loss based CNN model on source dataset as our baseline model, $f_{\theta_0}$, which acts as an initial estimate of non-linear mapping for target data.  $f_{\theta_0}$ is valuable for initialization due to existence of domain-invariant features across different domains \cite{Li2018AdaptationAR}.Thus, baseline initialization assists in faster convergence and better results.

\subsection{Cluster Assignment}
The next step is to assign each data point in unlabeled target dataset ($T$) to a cluster. Cluster assignment is equivalent to defining a class label for each unlabeled sample. Following Ding et al.\cite{feature_affinity} we use similarity between feature representation of a person image and cluster centers as a measure to decide which center the image belongs to. Cluster centers are initialized using standard \textit{k-means} clustering on features, and distance between an image and cluster center is measured by cosine similarity:
\begin{equation}
  sim(x_i \,,\, c_k) = \frac{f_\theta(x_i) \,.\, c_k}{\left| f_\theta(x_i) \right|\, \left|c_k \right|}
\end{equation}
where $f_{\theta}(x_i)$ represents the feature vector of sample image $x_i$ and $c_k$ denotes $kth$ cluster center out of $K$ total clusters. Now, we can assign a label $l$ to each data point as:
\begin{equation}
  l(x_i) = \underset{k}{argmax} \; sim(x_i \,,\, c_k)\,, \quad where \, k \in \left[1, K\right]
  \label{eq:assign_cluster}
\end{equation}
% This process can essentially be realized as a form of clustering process that assigns a pseudo-label to each person image.

\begin{figure*}[t]
    \centering
    \includegraphics[width=\textwidth]{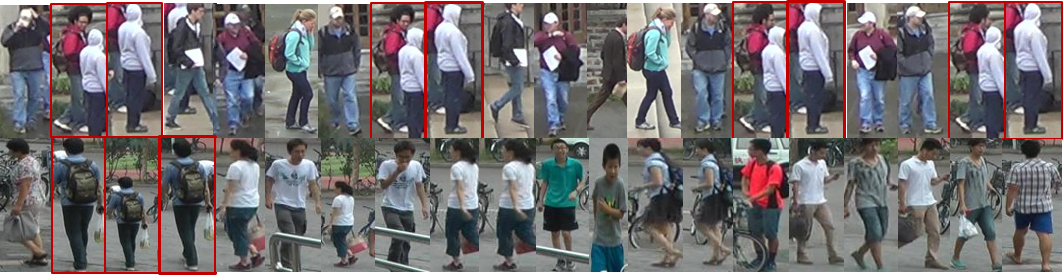}
    \caption{Top and bottom rows show image fragments sorted by Frame IDs from cameraID 2 of DukeMTMC-reID and Market-1501, respectively. Each fragment consists of images captured by one camera during a very short period. We can see that multiple images of one person exist in a fragment verifying their temporal closeness within a camera.}
    \label{fig:temp_imgs}
\end{figure*}

\subsection{Sample Selection}
\begin{figure}[t!]
    \centering
    \includegraphics[width= 7 cm]{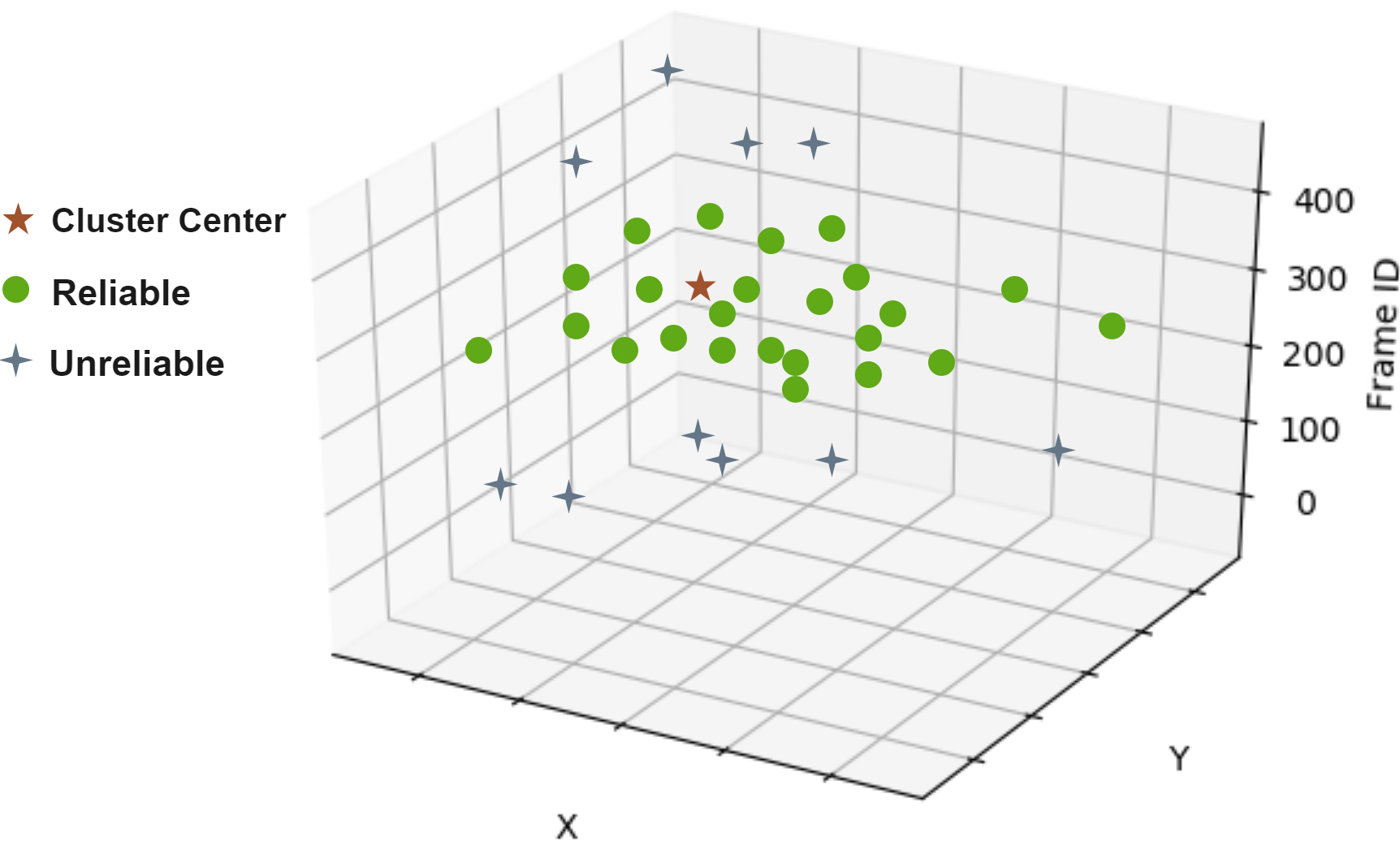}
    \caption{The figure shows data-points belonging to one camera in a single cluster along-with their Frame IDs. Data-points are marked reliable or unreliable based on the distance of their Frame IDs from that of Cluster Center}
    \label{fig:selection}
\end{figure}
Cluster assignment process yields unreliable pseudo-labels for target dataset due to existence of domain-specific features in source dataset used to train baseline model $f_{\theta_0}$, that do not exist in target dataset \cite{Li2018AdaptationAR}.  Fan et al \cite{Fan2018UnsupervisedPR} use cosine similarity of features with cluster-centers to filter out unreliable data points. However, this approach of choosing reliable samples has two issues. Firstly, during the first few iterations of training process, when the model has mostly seen source data only, a higher cosine similarity value does not necessarily translate to a higher confidence on pseudo-labels due to existence of domain specific features. Secondly, this method of choosing high confidence data points has an intrinsic problem of being uninformative because performance of re-id task is more sensitive to different pairs with minimal distance as opposed to similar pairs with minimal distance \cite{Hermans2017InDO}.  If all the selected data points are in close proximity to their respective cluster centers, then data points belonging to different cluster centers (i.e. differently labeled pairs) will be maximally distant from each other in feature space rendering them ineffective \cite{Hermans2017InDO}. Instead, we propose to exploit temporal continuity of images within a camera for sample selection. This proposition is based on two basic observations typical to a camera network \cite{Li2018UnsupervisedPR}. \textbf{(1)} Most people would appear in a camera view only once during a short period used for recording. \textbf{(2)}  Most people in a public scene would travel through a single camera view in a small common time. These observations imply that all images of a person taken by one camera would appear in a short continuous period, assuming no reappearance.  Figure \ref{fig:temp_imgs} verifies this assumption on two real-world datasets.

Specifically, the idea is that if two images within one camera belong to one identity, they should also be temporal neighbors. To put it another way, two images belonging to one camera with similar pseudo-labels must have Frame IDs close to each other. In order to select reliable samples, we define a temporary cluster center for the subset of images belonging to one camera in a particular cluster. Let $X_{jk}$ be all the images belonging to camera $j$ and cluster $k$. We define $x_{min_{jk}} \in \, X_{jk}$  such that $sim(f_{\theta}(x_{min_{jk}})\,,\,c_k) \geq sim(f_{\theta}(x_{jk})\,,\, c_k)$ where  $x_{jk} \in \, X_{jk}$ and $x_{jk} \neq x_{min_{jk}}$. We then define $c_{jk} = f_\theta(x_{min_{jk}})$  as a temporary cluster center for $X_{jk}$. Reliable samples are then selected as all those images in $X_{jk}$ whose Frame ID is within a threshold to the Frame ID of $x_{min_{jk}}$ as shown in Figure \ref{fig:selection}. Mathematically, selection function for an image $x_{jk}$ in camera $j$  and cluster $k$ is defined as the following indicator function:
\begin{equation}
  S(x_{jk}) = I(\left| fid(x_{min_{jk}}) - fid(x_{jk}) \right| \leq \lambda_{fr})
  \label{eq:selection}
\end{equation}
where $fid(.)$ gives the Frame ID, $I$ is an indicator function and $\lambda_{fr}$ is a positive integer for limiting the range of Frame IDs belonging to one identity . Note that this reliable sample selection procedure ensures that selected samples with similar pseudo-labels are close in feature space across-cameras and are neighbors temporally within-camera. 

\subsection{Fine-Tuning}
In order to fine-tune the CNN model over selected samples, we minimize a mini-batch based triplet loss to pull similarly labeled images closer in feature space while pushing differently labeled images away. Given $x_a$, $x_p$ and $x_n$ as anchor, positive and negative image with labels $l_a$, $l_p = l_a$ and $l_n \neq l_a$ respectively, triplet loss is computed as:

\begin{equation}
  L_{trip}(x_a, x_p, x_n) = \left[ D(x_a \,,\, x_p) - D(x_a \,,\, x_n) + m \right]_+
\end{equation}
where $m$ is a hard margin and  $D(. \,,\, .)$ defines the feature distance between two images which is fixed as Euclidean distance in this work. $[d]_+$ equals $d$ if $d > 0$ and zero otherwise. Overall triplet loss is then defined as:

\begin{equation}
  L_{trip} = \sum_{x_a: x_a,x_p,x_n \in B} L_{trip}(x_a, x_p, x_n)
\end{equation}
where $B$ denotes a mini-batch. Inspired by Hermans et al. \cite{Hermans2017InDO}, we perform Online Hard Mining (OHM) to generate triplets that results in faster convergence and better performance.

\subsection{Center Update}
Finally, cluster centers are updated using the features from fine-tuned CNN model such that centers are better representative of the target data. Specifically, for each cluster center $c_k$ we compute an update term as given by:
\begin{equation}
  \Delta c_k = \frac{I(l_i = k) . (c_k - f_\theta(x_i))}{\beta + \sum_{1=1}^N I(l_i = k)}
  \label{eq:delta}
\end{equation}
where $\beta$ is a small positive number to avoid division by zero. $l_i$ denotes the pseudo-label for image $x_i$, $N$ is the size of target data and $I(condition)$ is an indicator function. Cluster centers are updated by following simple equation:
\begin{equation}
  c_{k_{updated}} = c_k - \lambda_c \Delta c_k 
  \label{eq:center_update}
\end{equation}
where $\lambda_c$ is regularization parameter to control the change in cluster centers.  As discussed earlier, cluster update using Equation \ref{eq:center_update} provides a smooth transition from one set of cluster centers to the next by perturbing the original center values only by a small amount given by Equation \ref{eq:delta}

\begin{algorithm}[t]
    \SetAlgoLined
    \SetKwInOut{Input}{Input}
    \SetKwInOut{Output}{Output}
    \Input{Labelled source data $S$;
    Unlabeled target data  $T = \{x_i\}_{i=1}^N$;
    Number of iterations $N_{iter}$;
    Number of clusters $K$;
    Selection frame range $\lambda_{fr}$;
    Triplet margin $m$;
    Center update coefficient $\lambda_c$}
    \Output{Re-id model $f_{\theta}(.)$ for target data}
    Train a model $f_{\theta_0}$ on $S$\;
    k-means : initialize $\{c_k\}_{k=1}^K$\;
     \For{$iter = 1$ to $N_{iter}$}{
        extract features: $z_i = f_{\theta_{iter-1}(x_i)}$\;
        $l_i:$ assign clusters using Eq. \ref{eq:assign_cluster}\;
        $D_{iter} : $ Select reliable data using Eq. \ref{eq:selection}\;
        Fine-tune: $f_{\theta_{iter}} = $ Train model on $D_{iter}$\;
        Update Centers $\{c_k\}_{k=1}^K$ using Eq. \ref{eq:center_update}\;
 }
 \caption{Temporal Continuity Based Unsupervised Learning}
 \label{alg:tcul}
\end{algorithm}
\subsection{Overall Algorithm}
Algorithm \ref{alg:tcul} describes the overall training process for TCUL.  Apart from initialization step, algorithm performs multiple iterations over the remaining four steps. With each iteration, discriminative capability of the model is improved, thus resulting in better clusters with larger reliable sample size. A larger training sample with high confidence pseudo-labels in turn further enhances the discriminative ability of model. Hence, TCUL can be regarded as self-paced progressive learning algorithm.

\section{Experiments}
\label{sec:exp}
\subsection{Datasets and Implementation Details}
We evaluate our proposed method on three large-scale benchmark person re-id datasets, Market-1501 (Market) \cite{scalabale_reid} and DukeMTMC-reID (Duke) \cite{Ristani2016PerformanceMA,Zheng2017UnlabeledSG} and MSMT17 (MSMT) \cite{msmt17}. \ref{tab:datasets} shows the number of identities and images in training and test splits of each dataset.

We use ResNet-50 \cite{He2016DeepRL} as our backbone CNN model. During baseline model training on source dataset, we add a batch normalization layer after global average pooling layer followed by a rectified linear unit (ReLU). After ReLU layer, we insert two fully connected layer of size 2048 and 702 (depending upon number of identities in source training data), respectively. We train the baseline model using triplet loss and cross entropy loss. Adam optimizer with a learning rate of $1 \times 10^{-4}$ is used for optimization. Each training batch contains 16 identities with 8 images per identity resized to $256 \times 128$. The baseline model is trained for 150 epochs.

When training on unlabeled target data, last fully connected layer is removed. Triplet loss along-with stochastic gradient decent (SGD) is used for optimization. During each iteration, we perform data augmentation by randomly erasing and flipping the images. The model is fine-tuned for 20 epochs in one iteration. Other training parameters are as follows: number of identities per batch = 16, number of images per identity = 8, number of iterations = 10, learning rate = $5 \times 10^{-4}$, momentum = 0.9, m = 0.5, $\lambda_c = 0.5$, $\lambda_{fr} = 100$. Note that we use pseudo-labels generated during cluster assignment step as identities for sampling mini-batches. For evaluation purposes, we report rank-1 and 5 accuracy along with mean average precision (mAP) following Zheng et al. \cite{scalabale_reid} . All experiments employ single query.

\subsection{Comparison with State-of-the-art Methods}

% Please add the following required packages to your document preamble:
% \usepackage{multirow}
\begin{table}[t]
\caption{Dataset statistics}
\label{tab:datasets}
\begin{tabular}{|l|c|cc|cc|cc|}
\hline
\multirow{3}{*}{Datasets} & \multirow{3}{*}{\#Cams} & \multicolumn{2}{c|}{Training} & \multicolumn{4}{c|}{Test} \\ \cline{3-8} 
 &  & \multirow{2}{*}{\#IDs} & \multirow{2}{*}{\#images} & \multicolumn{2}{c|}{query} & \multicolumn{2}{c|}{gallery} \\ \cline{5-8} 
 &  &  &  & \#IDs & \#images & \#IDs & \#images \\ \hline
Market-1501 \cite{scalabale_reid} & 6 & 751 & 12,936 & 750 & 3,368 & 750 & 19,732 \\
DukeMTMC-reID \cite{Ristani2016PerformanceMA} & 8 & 702 & 16,522 & 702 & 2,228 & 1,110 & 17,661 \\
MSMT17 \cite{msmt17} & 15 & 1,041 & 32,621 & 3,060 & 11,659 & 3,060 & 82,161 \\ \hline
\end{tabular}
\end{table}

 % Please add the following required packages to your document preamble:
% \usepackage{multirow}
% \usepackage{graphicx}
\begin{table}[t!]
\centering
\caption{Comparison with state-of-the-art methods. A$\xrightarrow[]{}$B refers to A as source dataset and B as target dataset.}
\label{tab:results}
\resizebox{\textwidth}{!}{%
\begin{tabular}{|l|ccc|ccc|ccc|ccc|}
\hline
\multirow{2}{*}{Method} & \multicolumn{3}{c|}{Duke$\xrightarrow[]{}$Market} & \multicolumn{3}{c|}{Market$\xrightarrow[]{}$Duke} & \multicolumn{3}{c|}{MSMT$\xrightarrow[]{}$Market} & \multicolumn{3}{c|}{MSMT$\xrightarrow[]{}$Duke} \\ \cline{2-13} 
 & rank1 & rank5 & mAP & rank1 & rank5 & mAP & rank1 & rank5 & mAP & rank1 & rank5 & mAP \\ \hline
baseline & 42.8 & 57.2 & 18.8 & 26.5 & 41.5 & 11.8 & 47.7 & 66.5 & 24.2 & 48.5 & 65.1 & 29.2 \\
PUL \cite{Fan2018UnsupervisedPR} & 44.7 & 59.1 & 20.1 & 30.4 & 44.5 & 16.4 & - & - & - & - & - & - \\
TJ-AIDL \cite{Wang2018TransferableJA} & 58.2 & 74.8 & 26.5 & 44.3 & 59.6 & 23.0 & - & - & - & - & - & - \\
ARN \cite{Li2018AdaptationAR} & 70.3 & 80.4 & 39.4 & 60.2 & 73.9 & 33.4 & - & - & - & - & - & - \\
SPGAN + LMP \cite{Deng2018ImageImageDA} & 58.1 & 76.0 & 26.9 & 49.6 & 62.6 & 26.4 & - & - & - & - & - & - \\
TAUDL \cite{Li2018UnsupervisedPR} & 63.7 & - & 41.2 & 61.7 & - & 43.5 & - & - & - & - & - & - \\
SUL (ours) & 72.5 & 86.4 & 47.5 & 62.2 & 77.4 & 42.5 & 69.1 & 81.5 & 40.3 & 67.1 & 77.3 & 44.9 \\
TCUL (ours) & \textbf{75.8} & \textbf{89.0} & \textbf{51.7} & \textbf{64.0} & \textbf{79.2} & \textbf{45.0} & \textbf{72.1} & \textbf{86.3} & \textbf{44.5} & \textbf{69.8} & \textbf{81.9} & \textbf{48.6} \\ \hline
\end{tabular}%
}
\end{table}

 We compare our method (TCUL) with existing state-of-the-art unsupervised person re-id approaches on Market, Duke and MSMT datasets in Table \ref{tab:results}. Firstly, these results show that TCUL improves the baseline results by a large margin. Specifically, it achieves an improvement of $32.9\%$ and $33.2\%$ in mAP values on Market$\xrightarrow[]{}$Duke and Duke$\xrightarrow[]{}$Market, respectively. Similar improvements can also be observed for on MSMT$\xrightarrow[]{}$Duke and MSMT$\xrightarrow[]{}$Market \footnote[1]{MSMT is only used as source dataset since Frame ID information is not available.}. These large advancements verify that TCUL is an effective framework to utilize learned representations on source dataset for learning discriminative representation on target dataset. Secondly, we observe that our approach achieves a clear improvement on Market and Duke datasets when compared with all existing state-of-the-art methods. As mentioned earlier, TCUL outperforms PUL by a large margin due to superior clustering mechanism and a better selection approach which exploits temporal continuity.  Superiority of temporal continuity based selection will be further verified in the next subsection.

 %[mAP declines as we increase the value for $\lambda_{fr}$]
 %[mAP attains maximum value at K=750 which is closest value to actual number of clusters in both datasets]
\begin{figure*}[t!]
    \centering
    \subfloat[mAP declines as we increase the value for $\lambda_{fr}$]{\includegraphics[width=0.45\textwidth]{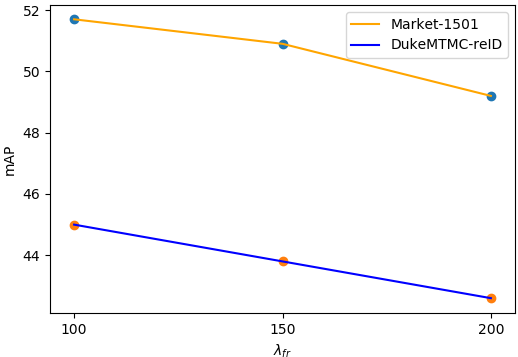}}
    \hfill
    \subfloat[mAP attains maximum value at K=750]{\includegraphics[width=0.45\textwidth]{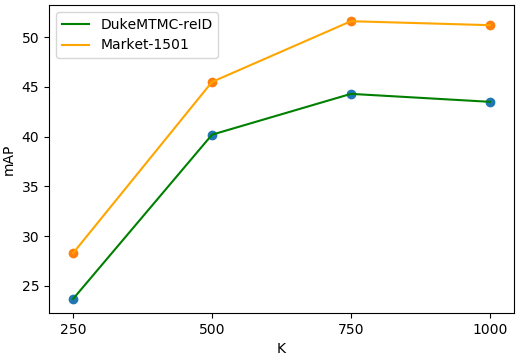}}
    \caption{Impact of parameter changes on TCUL}
    \label{fig:parmas_var}
\end{figure*}
 
 \subsection{Further Evaluation of TCUL}
\label{ss:eval}
In order to highlight the effect of temporal continuity based sample selection, we follow PUL \cite{Fan2018UnsupervisedPR} to select reliable samples by measuring similarity of each feature vector from respective cluster center. We employ the threshold value 0.85 for sample selection as used in PUL. We call this modified framework as Similarity-based Unsupervised Learning (SUL). Table \ref{tab:results} shows that performance of SUL has significantly reduced as compared to TCUL for all three datasets verifying the effectiveness of temporal continuity based reliable sample section. Table \ref{tab:results} also highlights the overall superiority of our framework, as SUL, which uses the same sample selection mechanism as PUL, still outperforms all existing methods.

Figure \ref{fig:parmas_var}a and \ref{fig:parmas_var}b show impact of parameter changes on mAP. As images belonging to one person in a camera should be temporally close to each other, increasing the value of $\lambda_{fr}$ above an optimal value should drop reliability of selected samples. Figure \ref{fig:parmas_var}a verifies this where increasing $\lambda_{fr}$ from 150 to 200 significantly lowers the mAP values for both Market and Duke datasets. These results also highlight the importance of reliable sample selection since removing it is equivalent to increasing the value of $\lambda_{fr}$ to maximum. Following the trends shown in Figure \ref{fig:parmas_var}a, we can deduce that performance of re-id model will further fall down if there is no sample selection step. Furthermore, Figure \ref{fig:parmas_var}b evaluates impact of parameter $K$ on mAP by varying it to 250, 500, 750 and 1000. As expected,  maximum performance is attained when $K=750$  which is  the nearest value to actual number of identities in each dataset (751 and 702).

\begin{figure*}
    \centering
    \subfloat{\includegraphics[width=0.5\textwidth]{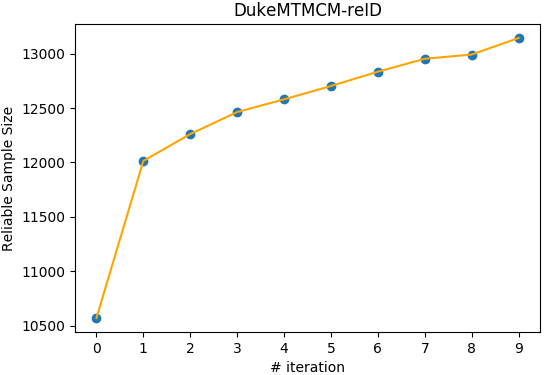}}
    \subfloat{\includegraphics[width=0.5\textwidth]{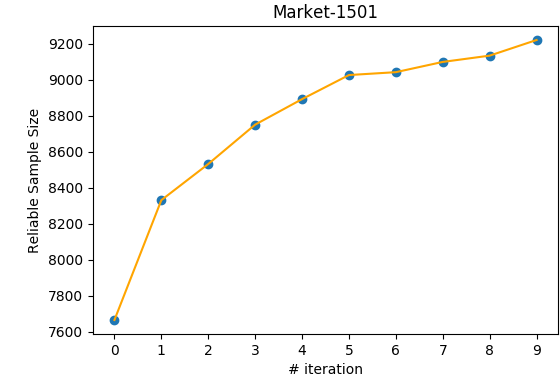}}
    \caption{Reliable sample size increases after each iteration}
    \label{fig:data_sample_size}
\end{figure*}

Fianlly, we also compute the variance in reliable samples size after each iteration. Ideally, as the training progresses, discriminative capability of the CNN model should increase. Consequently, we will have better cluster assignment producing more reliable samples. Figure \ref{fig:data_sample_size} shows that after each iteration, there is a steady growth in the size of reliable sample set for both datasets. Therefore, we can conclude that clustering and fine-tuning steps alternatively improve each other by generating pseudo-labels that are more reliable.

\section{Conclusion}
In this paper, we proposed a temporal continuity based unsupervised learning (TCUL) method for person re-id. It iterates between similarity-based clustering, reliable sample selection using temporal continuity of images within a camera and fine-tuning a CNN model on selected samples. Experimental results show that TCUL outperforms existing unsupervised person re-id methods. Furthermore, we also proved the significance of reliable sample selection via temporal continuity to achieve better discriminative ability for re-id task. Finally, we also utilized a pre-learned feature representation on source dataset for initialization; however, we did not actively exploit any state-of-the-art transfer-learning algorithms for representation transfer. This can be a promising future work to jointly utilize transfer-learning algorithms with temporal continuity of images within a camera to further enhance unsupervised person re-id.

%
% ---- Bibliography ----
%
% BibTeX users should specify bibliography style 'splncs04'.
% References will then be sorted and formatted in the correct style.
%
% \bibliographystyle{splncs04}
% \bibliography{mybibliography}
%

\bibliographystyle{splncs04}
\bibliography{sample-base}

\begin{thebibliography}{10}
\providecommand{\url}[1]{\texttt{#1}}
\providecommand{\urlprefix}{URL }
\providecommand{\doi}[1]{https://doi.org/#1}

\bibitem{Chen2018DeepAL}
Chen, Y., Zhu, X., Gong, S.: Deep association learning for unsupervised video
  person re-identification. In: BMVC (2018)

\bibitem{Cheng2016PersonRB}
Cheng, D., Gong, Y., Zhou, S., Wang, J., Zheng, N.: Person re-identification by
  multi-channel parts-based cnn with improved triplet loss function. CVPR
  (2016)

\bibitem{Deng2018ImageImageDA}
Deng, W., Zheng, L., Ye, Q., Kang, G., Yang, Y., Jiao, J.: Image-image domain
  adaptation with preserved self-similarity and domain-dissimilarity for person
  re-identification. In: CVPR 2018 (2018)

\bibitem{Ding2017LetFD}
Ding, G., Khan, S.H., Tang, Z., Porikli, F.M.: Let features decide for
  themselves: Feature mask network for person re-identification. CoRR
  \textbf{abs/1711.07155} (2017)

\bibitem{feature_affinity}
Ding, G., Zhang, S., Khan, S., Tang, Z., Zhang, J., Porikli, F.: Feature
  affinity based pseudo labeling for semi-supervised person re-identification.
  IEEE Transactions on Multimedia  (2019)

\bibitem{Fan2018UnsupervisedPR}
Fan, H., Zheng, L., Yang, Y.: Unsupervised person re-identification: Clustering
  and fine-tuning. TOMCCAP  \textbf{14},  83:1--83:18 (2018)

\bibitem{He2016DeepRL}
He, K., Zhang, X., Ren, S., Sun, J.: Deep residual learning for image
  recognition. CVPR  (2016)

\bibitem{Hermans2017InDO}
Hermans, A., Beyer, L., Leibe, B.: In defense of the triplet loss for person
  re-identification. In: CoRR (2014)

\bibitem{Hirzer2012RelaxedPL}
Hirzer, M., Roth, P.M., K{\"o}stinger, M., Bischof, H.: Relaxed pairwise
  learned metric for person re-identification. In: ECCV (2012)

\bibitem{Khan2016UnsupervisedDA}
Khan, F.M., Br{\'e}mond, F.: Unsupervised data association for metric learning
  in the context of multi-shot person re-identification. 2016 13th IEEE
  International Conference on Advanced Video and Signal Based Surveillance
  (AVSS)  (2016)

\bibitem{Kodirov2016PersonRB}
Kodirov, E., Xiang, T., Fu, Z.Y., Gong, S.: Person re-identification by
  unsupervised l1 graph learning. In: ECCV (2016)

\bibitem{Kodirov2015DictionaryLW}
Kodirov, E., Xiang, T., Gong, S.: Dictionary learning with iterative laplacian
  regularisation for unsupervised person re-identification. In: BMVC (2015)

\bibitem{Li2018UnsupervisedPR}
Li, M., Zhu, X., Gong, S.: Unsupervised person re-identification by deep
  learning tracklet association. In: ECCV (2018)

\bibitem{filter_pair}
Li, W., Zhao, R., Xiao, T., Wang, X.: Deepreid: Deep filter pairing neural
  network for person re-identification. In: CVPR (2014)

\bibitem{Li2018HarmoniousAN}
Li, W., Zhu, X., Gong, S.: Harmonious attention network for person
  re-identification. CVPR  (2018)

\bibitem{Li2018AdaptationAR}
Li, Y.J., Yang, F.E., Liu, Y.C., Yeh, Y.Y., Du, X., Wang, Y.C.F.: Adaptation
  and re-identification network: An unsupervised deep transfer learning
  approach to person re-identification. CVPRW  (2018)

\bibitem{Lisanti2015PersonRB}
Lisanti, G., Masi, I., Bagdanov, A.D., Bimbo, A.D.: Person re-identification by
  iterative re-weighted sparse ranking. IEEE Transactions on Pattern Analysis
  and Machine Intelligence  \textbf{37},  1629--1642 (2015)

\bibitem{Lisanti2014MatchingPA}
Lisanti, G., Masi, I., Bimbo, A.D.: Matching people across camera views using
  kernel canonical correlation analysis. In: ICDSC (2014)

\bibitem{Ma2017PersonRB}
Ma, X., Zhu, X., Gong, S., Xie, X., Hu, J., Lam, K.M., Zhong, Y.: Person
  re-identification by unsupervised video matching. Pattern Recognition
  \textbf{65} (2017)

\bibitem{Paisitkriangkrai2015LearningTR}
Paisitkriangkrai, S., Shen, C., van~den Hengel, A.: Learning to rank in person
  re-identification with metric ensembles. CVPR  (2015)

\bibitem{Peng2016UnsupervisedCT}
Peng, P., Xiang, T., Wang, Y., Pontil, M., Gong, S., Huang, T., Tian, Y.:
  Unsupervised cross-dataset transfer learning for person re-identification.
  CVPR  (2016)

\bibitem{Ristani2016PerformanceMA}
Ristani, E., Solera, F., Zou, R.S., Cucchiara, R., Tomasi, C.: Performance
  measures and a data set for multi-target, multi-camera tracking. In: ECCV
  Workshops (2016)

\bibitem{Su2017PoseDrivenDC}
Su, C., Li, J., Zhang, S., Xing, J., Gao, W., Tian, Q.: Pose-driven deep
  convolutional model for person re-identification. ICCV  (2017)

\bibitem{Wang2018TransferableJA}
Wang, J., Zhu, X., Gong, S., Li, W.: Transferable joint attribute-identity deep
  learning for unsupervised person re-identification. CVPR  (2018)

\bibitem{Wei2018PersonTG}
Wei, L., Zhang, S., Gao, W., Tian, Q.: Person transfer gan to bridge domain gap
  for person re-identification. CVPR  (2018)

\bibitem{msmt17}
Wei, L., Zhang, S., Gao, W., Tian, Q.: Person trasfer gan to bridge domain gap
  for person re-identification. In: CVPR (2018)

\bibitem{Xiao2016LearningDF}
Xiao, T., Li, H., Ouyang, W., Wang, X.: Learning deep feature representations
  with domain guided dropout for person re-identification. CVPR  (2016)

\bibitem{Ye2017DynamicLG}
Ye, M., Ma, A.J., Zheng, L., Li, J., Yuen, P.C.: Dynamic label graph matching
  for unsupervised video re-identification. ICCV  (2017)

\bibitem{dml_reid}
Yi, D., Lei, Z., Liao, S., Li, S.Z.: Deep metric learning for person
  re-identification. In: ICPR (Aug 2014)

\bibitem{scalabale_reid}
Zheng, L., Shen, L., Tian, L., Wang, S., Wang, J., Tian, Q.: Scalable person
  re-identification: A benchmark. In: ICCV (2015)

\bibitem{zheng2016person}
Zheng, L., Yang, Y., Hauptmann, A.G.: Person re-identification: Past, present
  and future. CoRR  \textbf{abs/1610.02984} (2016)

\bibitem{Zheng2017ADL}
Zheng, Z., Zheng, L., Yang, Y.: A discriminatively learned cnn embedding for
  person reidentification. TOMCCAP  \textbf{14},  13:1--13:20 (2017)

\bibitem{Zheng2017UnlabeledSG}
Zheng, Z., Zheng, L., Yang, Y.: Unlabeled samples generated by gan improve the
  person re-identification baseline in vitro. ICCV  (2017)

\end{thebibliography}

\end{document}